\documentclass[runningheads]{llncs}
\usepackage{graphicx}
\usepackage{amsmath,amssymb} 
\usepackage{color}
\usepackage[ruled]{algorithm2e}
\usepackage[width=122mm,left=12mm,paperwidth=146mm,height=193mm,top=12mm,paperheight=217mm]{geometry}
\begin{document}
\pagestyle{headings}
\mainmatter

\title{Autonomously and Simultaneously Refining Deep Neural Network Parameters by Generative Adversarial Networks} 

\titlerunning{Autonomously and Simultaneously Refining DNN Parameters by GANs}

\authorrunning{}

\author{Yantao Lu, Burak Kakillioglu, and Senem Velipasalar}


\institute{Electrical Engineering and Computer Science,\\
	Syracuse University\\
	\email{ \{ylu25,bkakilli,svelipas\}@syr.edu}
}

\maketitle

\begin{abstract}
The choice of parameters, and the design of the network architecture are important factors affecting the performance of deep neural networks. However, there has not been much work on developing an established and systematic way of building the structure and choosing the parameters of a neural network, and this task heavily depends on trial and error and empirical results. Considering that there are many design and parameter choices, such as the number of neurons in each layer, the type of activation function, the choice of using drop out or not, it is very hard to cover every configuration, and find the optimal structure. In this paper, we propose a novel and systematic method that autonomously and simultaneously optimizes multiple parameters of any given deep neural network by using a generative adversarial network (GAN). In our proposed approach, two different models compete and improve each other progressively with a GAN-based strategy. Our proposed approach can be used to autonomously refine the parameters, and improve the accuracy of different deep neural network architectures. Without loss of generality, the proposed method has been tested with three different neural network architectures, and three very different datasets and applications. The results show that the presented approach can simultaneously and successfully optimize multiple neural network parameters, and achieve increased accuracy in all three scenarios.
\end{abstract}

\section{Introduction}
Deep learning-based techniques have found widespread use in machine learning. Even before the convolutional approaches becoming popular, simple multi-layer perceptron neural networks (MLPNN) had been widely used for classification tasks for several reasons. First, they are very easy to construct and run since each layer is represented and operated as a single matrix multiplication. Second, a neural network can become a complex non-linear mapping between input and output by the introduction of non-linear activation functions after each layer. Regardless of how large the input size (i.e. the number of features) or the output size (i.e. the number of classes) are, a neural network can discover relations between them when the network is sufficiently large; and enough samples, which cover the problem domain as much as possible, are provided during training.

Although a MLPNN is successful to form a complex non-linear relationship between the feature space and the output class space, it lacks the ability of discovering features by itself. Until recent years, the classical approach was to provide either the input data directly or some high-level descriptors, which are extracted by applying some algorithm on the input data, as the source of features. Using the raw input as features does not guarantee to yield any satisfactory mappings and the latter approach would require the investigation of multiple hand-crafted descriptor extraction algorithms for different applications. The introduction of convolutional layers in neural networks removed the necessity of having prior feature extractors, which are not easy to craft for different applications. Convolutional layers are designed to extract features directly from the input. Since they have been proven to be successful feature extractors and thanks to much faster computation of the operations of convolutional neural networks (CNNs) on specialized processors such as GPUs, the use of CNNs exploded recently. After Krizhevksy et al.~\cite{krizhevsky2012imagenet} achieved a significant increase in the classification accuracy on the ImageNet Large Scale Visual Recognition Challenge (ILSVRC)~\cite{deng2009imagenet} in 2012, many others followed their approach, by creating different architectures and applying them to numerous applications in many different domains.

It is well-known that the training of deep learning methods requires large amounts of data, and they usually perform better when training data size is increased. However, for some applications, it is not always possible to obtain more data when the dataset at hand is not large enough. In many cases, even though the raw data can be collected easily, the labeling or annotation of the data is difficult, expensive and time consuming. Successors of ~\cite{krizhevsky2012imagenet} yielded better accuracies with less number of parameters on the same benchmark with some architectural modifications using the same building blocks. This shows that the choice of parameters, and the design of the architecture are important factors affecting the performance. In fact, the design of a CNN model is very important to achieve better results, and many researchers have been working hard to find better CNN architectures \cite{simonyan2014very,zeiler2014visualizing,lin2013network,szegedy2015going,szegedy2016rethinking,he2016deep,ren2015faster} to achieve higher accuracy.

However, there has not been much work on developing an established and systematic way of building the structure of a neural network, and this task heavily depends on trial and error, empirical results, and the designer's experience. Considering that there are many design and parameter choices, such as the number of layers, number of neurons in each layer, number of filters at each layer, the type activation function, the choice of using drop out or not and so on, it is not possible to cover every possibility, and it is very hard to find the optimal structure. In fact, often times some common settings are used without even trying different ones. Moreover, the hyper-parameters in training phase also play important role on how well the model will perform. Likewise, these parameters are also tuned manually in an empirical way most of the time.

In this work, we focus on optimizing the network architecture and training parameters for any given neural network model. We propose a novel and systematic way, which employs generative adversarial networks (GANs) to find the optimal structure and parameters.

\subsection{Related Work}
There have been works focusing on optimizing neural network architectures. Most of the proposed approaches are based on the genetic algorithms (GA), or evolutionary algorithms, which are heuristic search algorithms. Benardos and Vosniakos~\cite{benardos2007optimizing} proposed a methodology for determining the best neural network architecture based on the use of a genetic algorithm and a criterion that quantifies the performance and the complexity of a network. In their work, they focus on optimizing four architecture decisions, which are the number of layers, the number of neurons in each layer, the activation function in each layer, and the optimization function. Islam et al.~\cite{islam2014optimization} also employ a genetic algorithm for finding the optimal number of neurons in the input and hidden layers. They apply their approach to power load prediction task and report better results than a manually designed neural network. However, their approach is used to optimize only the number of neurons for input and hidden layers, and optimization of other important design decisions such as the number of layers or type of activation functions are not discussed. Stanley and Miikkulainen~\cite{stanley2002evolving} presented the NEAT algorithm for optimizing neural networks by evolving topologies and weights of relatively small recurrent networks. In a recent work, Miikkulainen et al.~\cite{miikkulainen2017evolving} proposed CoDeepNEAT algorithm for optimizing deep learning architectures through evolution by extending existing neuroevolution methods to topology, components and hyperparameters. Ritchie et al.~\cite{ritchie2003optimizationof} proposed a method to automate neural network architecture design process for a given dataset by using genetic programming. The genetic algorithm-based optimization uses a given set of blueprints and models, i.e. it performs a finite search over a discrete set of candidates. Thus, genetic algorithms, in general, cannot generate unseen configurations, and they can only make a combination of preset parameters.

Apart from the genetic algorithms, Bergstra and Bengio~\cite{bergstra2012random} have proposed random search for hyper-parameter optimization, and stated that randomly chosen trials are more efficient for hyper-parameter optimization than trials on a grid. Yan and Zhang~\cite{jin2016neural} optimized architectures' width and height with growing running time budget through submodularity and suparmodularity.

Generative Adversarial Networks (GANs)~\cite{goodfellow2014generative} are one of the important milestones in deep learning research. In contrast to CNNs, which extract rich and dense representations of the source domain, and may eventually map source instances into some classes, GANs generate instances of the source domain from small  noise. They employ deconvolution operators, or transposed convolutions, to generate N-D instances from 1-D noise. GAN's power comes from the competition with the discriminator, which decides whether the generated instance belongs to the source domain. Discriminator acts like the police who is trying to intercept counterfeit money, where in this case the generator is the counterfeiter. Generator and discriminator are trained together until discriminator cannot distinguish the generated instances from the instances in the source domain. GANs have been adapted in many applications~\cite{gatys2015neural,radford2015unsupervised,isola2017image,zhu2017unpaired}.

In our proposed approach, two different models compete and improve each other progressively with a GAN-based strategy. Our proposed approach can be used to autonomously refine the parameters, and improve the accuracy of different deep neural networks. For this work, we have tested the performance of our approach on three different neural network structures covering Long Short Term Memory (LSTM) networks and 3D CNNs, and different applications. Without loss of generality, we have chosen simpler network structures (not necessarily very deep ones) to optimize in order to show that the performance improvement is obtained not because of increasing number of layers, but instead thanks to better refinement and optimization of the network parameters.

The rest of this paper is organized as follows: The proposed method is described in Sec. \ref{sec:proposed}. The results showing the accuracy improvement on three different neural network structures with three different datasets/applications are presented in Sec. \ref{sec:exp}, and the paper is concluded in Sec. \ref{sec:concl}.

\section{Proposed Method}\label{sec:proposed}
\subsection{Overview}
\begin{figure}[t]
	\centering
	\includegraphics[width=.8\textwidth]{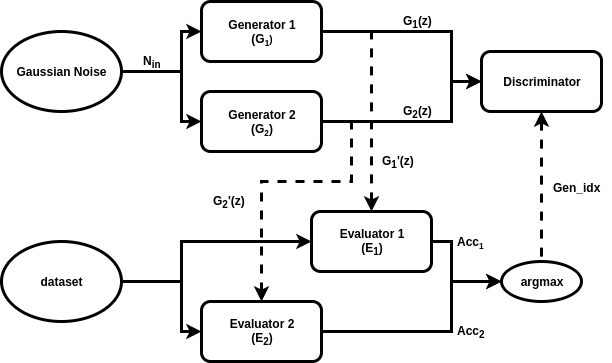}
	\caption{Proposed GAN-based network for refining deep neural network parameters.}
	\label{fig:network}
\end{figure}

We propose a novel and systematic way, which employs generative adversarial networks (GANs) to find the optimal network structure and parameters. The proposed GAN-based network for refining different neural network parameters is shown in Fig.~\ref{fig:network}. It is composed of a generative part, an evaluation part and a discriminator. There are two generators ($G_1$ and $G_2$), two evaluators ($E_1$ and $E_2$), and one discriminator ($D$). The input to the two generators is Gaussian noise $z \sim p_{noise}(z)$. On the other hand, the input to the evaluators is the training data $x \sim p_{data}(x)$.

As will be discussed in more detail below, the generators $G_1$ and $G_2$ have the same network structure. From input noise $p_{noise}(z)$, $G_1$ and $G_2$ generate the input network parameters $G_1(z)$ and $G_2(z)$ to be used and evaluated by $E_1$ and $E_2$, respectively. $E_1$ and $E_2$ have the structure of the neural network whose parameters are being optimized or refined. They calculate the classification accuracy on the training data $x$. $E_i(x,G_i(z)), i\epsilon\{1,2\}$ represents the classification accuracy obtained by the evaluator $E_i$ when the parameters $G_i(z)$ are used. The generator resulting in higher accuracy is marked as more accurate generator $G_a$, and the other generator is marked as $G_b$, where $a\in {1,2}, b=!a$.

We define the discriminator $D$ as a network, which is used for binary classification between better generator and worse generator. $G_1(z)$ and $G_2(z)$ are fed into the discriminator $D$, and the ground truth label about which is the better generator comes from the evaluators. The discriminator $D$ provides the gradients to train the worse performing generator.

The details of the proposed method are described in Sec. \ref{ssec:details}, and the pseudo code is provided in Algorithm 1.

\subsection{Details of the proposed network}\label{ssec:details}
\begin{figure}[hbt!]
	\centering
	\begin{minipage}{.5\textwidth}
		\centering
		\includegraphics[width=.4\linewidth]{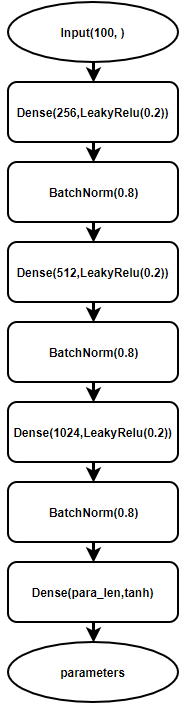}
		\caption{Generator network}
		\label{fig:generator}
	\end{minipage}%
	\begin{minipage}{.5\textwidth}
		\centering
		\includegraphics[width=.5\linewidth]{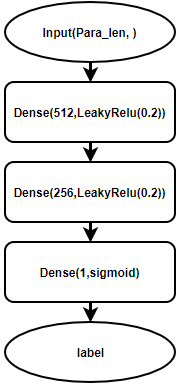}
		\caption{Discriminator network}
		\label{fig:discriminator}
	\end{minipage}
\end{figure}

\noindent \textbf{2.2.1 Generative part:} \\
\noindent The two generators $G_1$ and $G_2$ have the same neural network structure shown in Fig.~\ref{fig:generator}. Their input is a Gaussian noise vector $z$, and their outputs are $G_1(z)$ and $G_2(z)$. As seen in Fig.~\ref{fig:generator}, generators are composed of fully connected layers with leaky relu activations. At the output layer, $tanh$ is employed so that $G_i^j(z) \in (-1,1)$, where $j \in \{1,2,...,length(G_i(z))\}$ and $i \in \{1,2\}$. Then, the range of $G_i(z)$ is changed from $(-1,1)$ to $(pm_{min}^j,pm_{max}^j)$ by using
\begin{equation} \label{eqn:range}
G_i(z)'=[G_i(z)\times \frac{pm_{max}-pm_{min}}{2} + \frac{pm_{max}+pm_{min}}{2}].
\end{equation}
In (\ref{eqn:range}), $pm_{max}$ and $pm_{min}$ are preset maxima and minima values, which are defined empirically based on values that a certain parameter can take, so that the value of the refined parameters can only change between $pm_{max}$ and $pm_{min}$. The re-scaled values $G_1(z)'$ and $G_2(z)'$ are then used as parameters of evaluator networks. The length of $G_i(z)$ is determined by the number of network parameters that are refined, and is set at the generator network's last fully connected layer.

Generators are trained/improved by the discriminator, which is a binary classifier used to differentiate the results from generator outputs $G_1(z)$ and $G_2(z)$. Labels ``a" and ``b" represent the generators with higher accuracy and lower accuracy results, respectively. The generator, which has the worse performance and is labeled by ``b", is trained by stochastic gradient descent (SGD) from the discriminator to minimize $log(1-D(G_b(z))$ by using
\begin{equation}
\bigtriangledown_{G_b} \frac{1}{m}\displaystyle\sum_{j=1}^{m} log(1-D(G_b(z^{(j)}))),
\end{equation}
where $m$ is the number of epochs.

When $G_a'(z)$ becomes equal to $G_b'(z)$ for two consecutive iterations, the weights of $G_b$ will be re-initialized to default random values. The purpose of this step is to prevent the optimization stopping at a local maxima and also prevent the vanishing tanh gradient problem.
\\

\noindent \textbf{2.2.2 Evaluation part:} \\
\noindent As mentioned above, one of the strengths of the proposed approach is that it can be used to refine/optimize parameters of different deep neural network structures. In other words, the evaluator networks have the same structure as the neural network whose parameters are being optimized or refined. As will be shown in Sec. \ref{sec:exp}, we have tested the proposed approach with three different network structures, and different sets of parameters.

Evaluator networks are built by using the parameters $G_1(z)'$ and $G_2(z)'$ provided by the generators. The training data $x \sim p_{data}(x)$ is used to evaluate these network models. We employ an early stopping criteria. More specifically, if no improvement is observed in $c$ epoches, the training is stopped.

We then obtain the accuracies $acc_i=\mathbb{E}_{x\sim p_{data}(x)}E_i(x)$, $i=\{1,2\}$. Let $a$ be the value of $i$ resulting in higher accuracy, and $b=!a$. Then ``a" is used as the ground truth label for the discriminator, which marks the generator with better parameters, and trains the worse generator $G_b$.\\

\noindent \textbf{2.2.3 Discriminator:} \\
\noindent We define the discriminator $D$ as a network, whose output is a scalar softmax output, which is used for binary classification between better generator and worse generator. $G_1(z)$ and $G_2(z)$ are fed into the discriminator $D$, and the ground truth label about which is better generator comes from the evaluators. Let $D(G(z))$ represent the probability that $G(z)$ came from the more accurate generator $G_a$ rather than $G_b$. We train $D$ to maximize the probability of assigning the correct label to the outputs $G_1(z)$ and $G_2(z)$ of both generators. Moreover, we simultaneously train the worse generator $G_b$ to minimize $log(1-D(G_b(z))$. The whole process can be expressed by:
\begin{equation} \label{eqn:train_generator}
min_{G_a}max_D \mathbb{E}_{z\sim p_z(z)}(log(D(G_a(z)))+log(1-D(G_b(z)))),
\end{equation}
where, $a=argmax_{i=\{1,2\}} (\mathbb{E}_{x\sim p_{data}(x)}E_i(x))$, $b=!a$.

The pseudo-code for the entire process is provided in Algorithm 1.

\begin{algorithm}[htb]
	\caption{algorithm.}
	\While{in the iterations}{
		Generate $m\times2$ noise samples $\{Z_1^{(1)},Z_1^{(2)},...,Z_1^{(m)}\}$ and $\{Z_2^{(1)},Z_2^{(2)},...,Z_2^{(m)}\}$ from Gaussian white noise
		
		\While{j in range(m)}{
			Build evaluators $E_1^{(j)}$ and $E_2^{(j)}$ based on parameters from $G_1(Z_1^{(j)})$ and $G_2(Z_2^{(j)})$
			
			Calculate $acc_1^j$ and $acc_2^j$ from $\mathbb{E}_{x\sim p_{data}(x)}E_i(x)$
			
			End if no accuracy improvement after $c$ epoches
		}
		Calculate mean value $acc_i=(1/m)\sum_{j=1}^m acc_i^j$, $i=\{1,2\}$
		
		Find $G_a$ as $G_{argmax(acc_1,acc_2)}$ and $G_b$ as the other one.
		
		Update Discriminator by SGD:
		$\bigtriangledown_D \frac{1}{m}\displaystyle\sum_{j=1}^{m} (log(D(G_a(z^{(j)})))+log(1-D(G_b(z^{(j)}))))$
		
		Update Generator $G_b$ by SGD:
		$\bigtriangledown_{G_b} \frac{1}{m}\displaystyle\sum_{j=1}^{m} log(1-D(G_b(z^{(j)})))$
	}
	\label{alg:algorithm}
\end{algorithm}

\vspace{-0.3cm}
\section{Experimental Results} \label{sec:exp}
In order to show the promise of the proposed approach in autonomously and simultaneously refining multiple deep neural network parameters, we tested its performance, without loss of generality, with three different neural network structures, and different training data types and applications. Below, we describe the details of each scenario. The network architectures, whose parameters are optimized, are shown in Figures \ref{fig:ModelNet_evaluator}, \ref{fig:LSTM_evaluator} and \ref{fig:NLP}. In these figures, the parameters that are being refined/optimized are highlighted in red. 

\subsection{Experiments with ModelNet}
\begin{figure}[hbt!]
	\centering
	\begin{minipage}{.19\textwidth}
		\centering
		\includegraphics[width=1\linewidth]{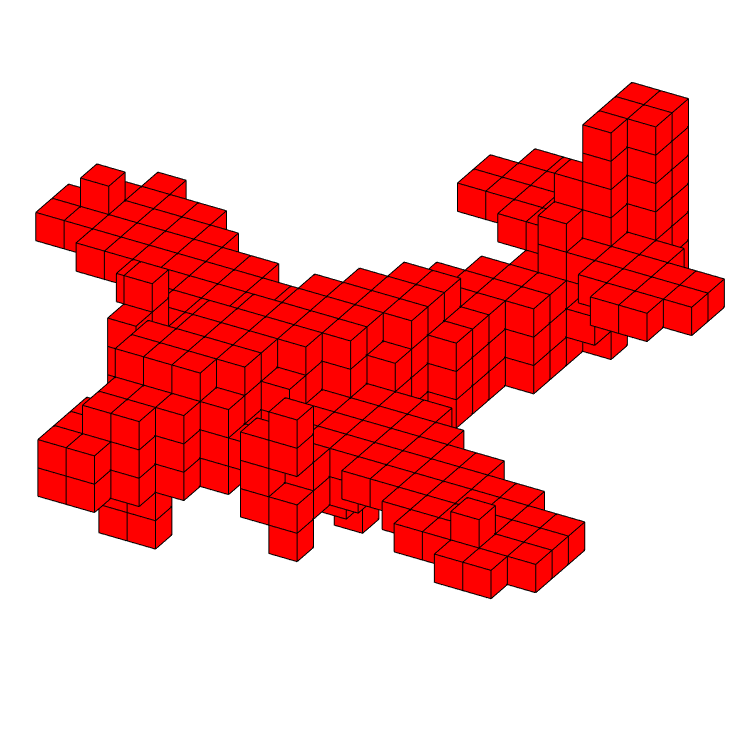}
	\end{minipage}%
	\begin{minipage}{.19\textwidth}
		\centering
		\includegraphics[width=1\linewidth]{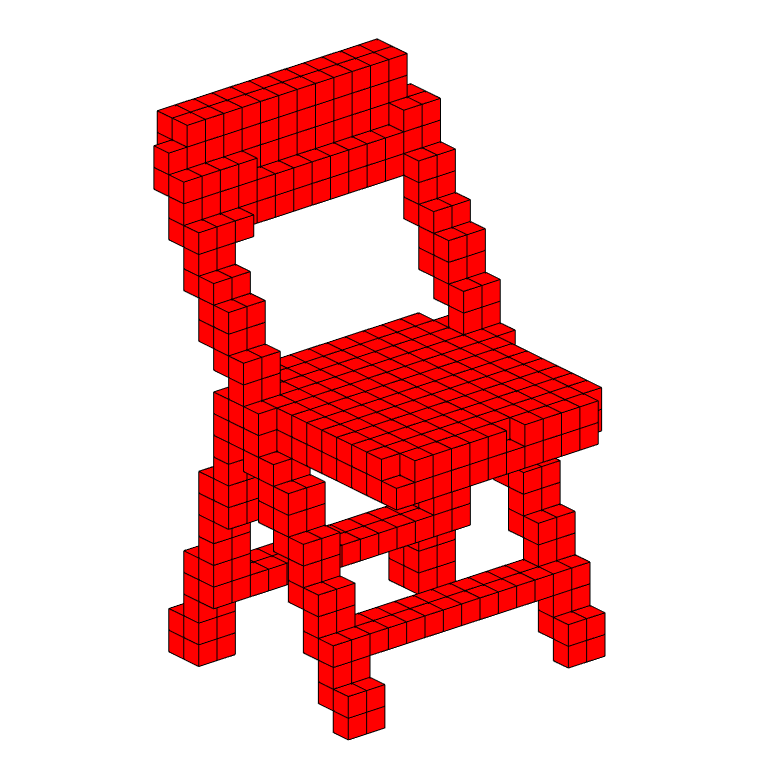}
	\end{minipage}
	\begin{minipage}{.19\textwidth}
		\centering
		\includegraphics[width=1\linewidth]{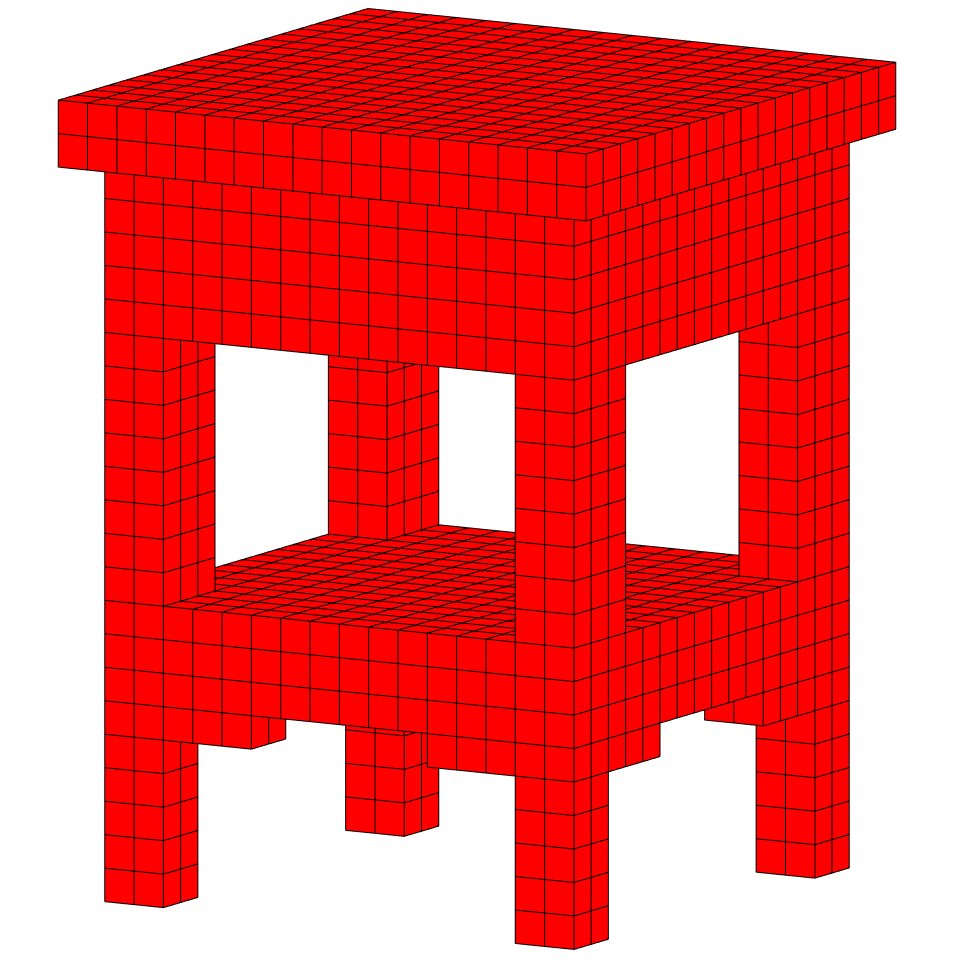}
	\end{minipage}
	\begin{minipage}{.19\textwidth}
		\centering
		\includegraphics[width=1\linewidth]{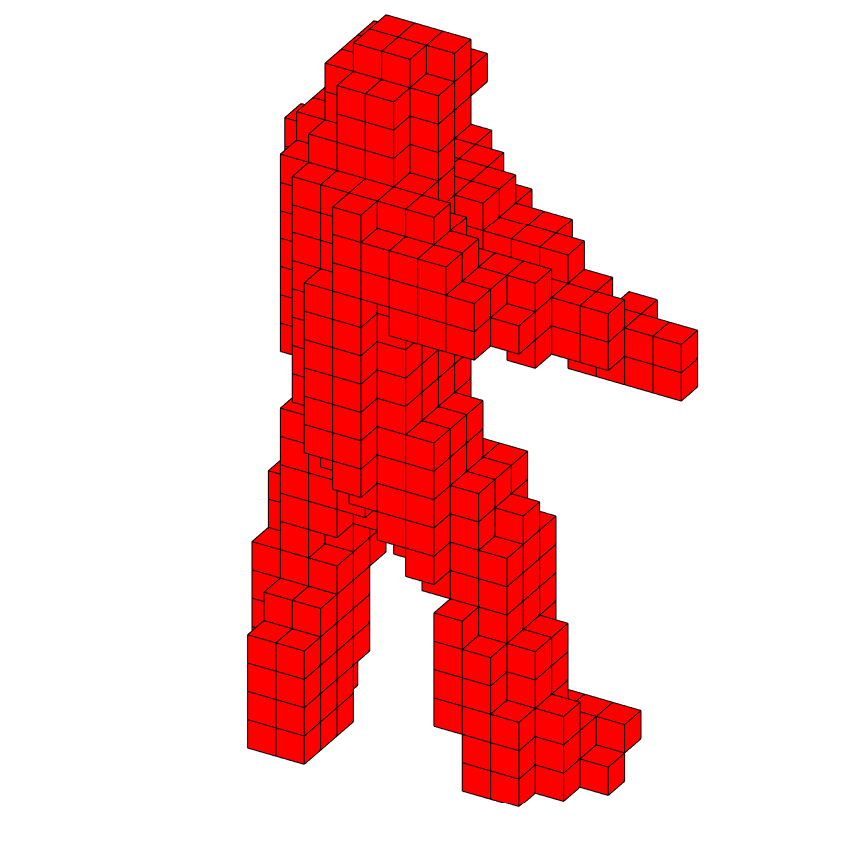}
	\end{minipage}
	\begin{minipage}{.19\textwidth}
		\centering
		\includegraphics[width=1\linewidth]{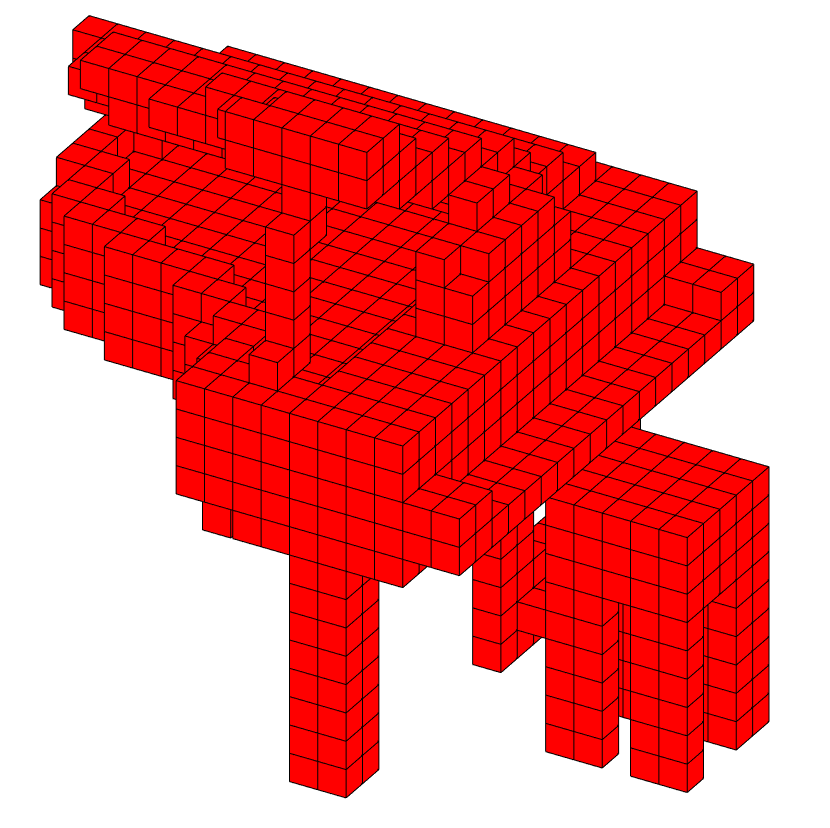}
	\end{minipage}
	\caption{Sample voxelized objects from ModelNet40 dataset.}
	\label{fig:exp_modelnet}
\end{figure}

We applied the proposed approach on a 3D convolutional network by using the ModelNet40 dataset~\cite{wu20153d}. ModelNet is a dataset of 3D point clouds. The goal is to perform shape classification over 40 shape classes. Some example voxelized objects from the ModelNet40 dataset are shown in Fig.~\ref{fig:exp_modelnet}.

\begin{figure}[hbt!]
	\centering
	\begin{minipage}{.29\textwidth}
		\centering
		\includegraphics[width=.97\linewidth]{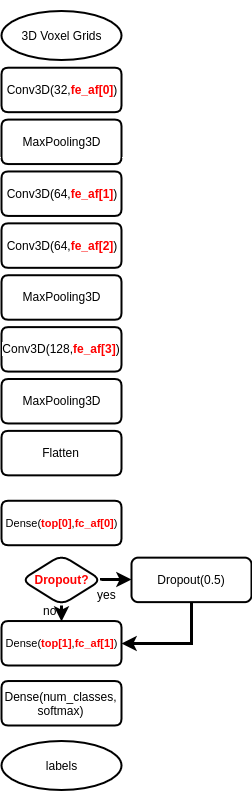}
		\caption{Evaluator network for shape classification on ModelNet}
		\label{fig:ModelNet_evaluator}
	\end{minipage}%
	\begin{minipage}{.38\textwidth}
		\centering
		\includegraphics[width=.95\linewidth]{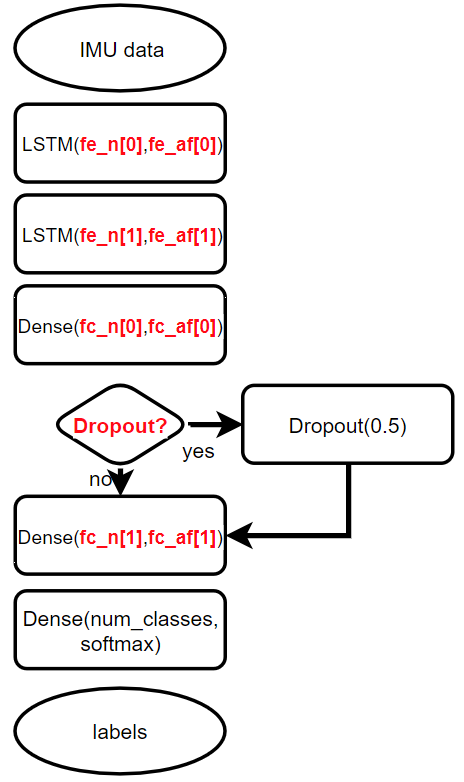}
		\caption{Evaluator network for activity classification}
		\label{fig:LSTM_evaluator}
	\end{minipage}
	\begin{minipage}{.32\textwidth}
		\centering
		\includegraphics[width=.95\linewidth]{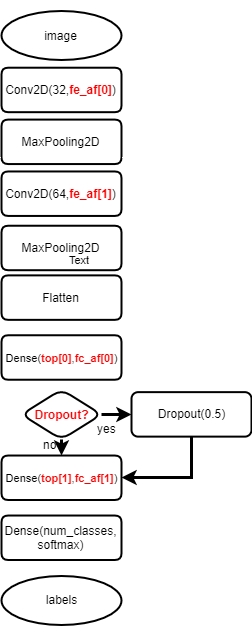}
		\caption{Evaluator network for character recognition}
		\label{fig:NLP}
	\end{minipage}
\end{figure}

The 3D CNN model shown in Fig.~\ref{fig:ModelNet_evaluator} is used for evaluators. The output of each generator is a 9-dimensional vector which is composed of different parameter settings. More specifically, two of the parameters are the number of neurons for two fully connected layers. Six of the parameters indicate the choice of activation function for fully connected and convolutional layers from (`Sigmoid', `Relu', `Linear', `Tanh') functions. One of the nine parameters is a flag indicating whether to add a dropout layer between fully connected layers. In this case, $pm_{max}$ and $pm_{min}$ are set to be: [4000, 4000, 4, 4, 4, 4, 4, 4, 1] and [1, 1, 0, 0, 0, 0, 0, 0, 0], respectively. Selecting the number of neurons is a regression problem and choosing the activation function is a classification problem. In other words, for choosing the activation function, the $tanh$ output is put into bins, and the corresponding function is selected.

\begin{figure}[h!]
	\centering
	\centerline{\includegraphics[width=.7\textwidth]{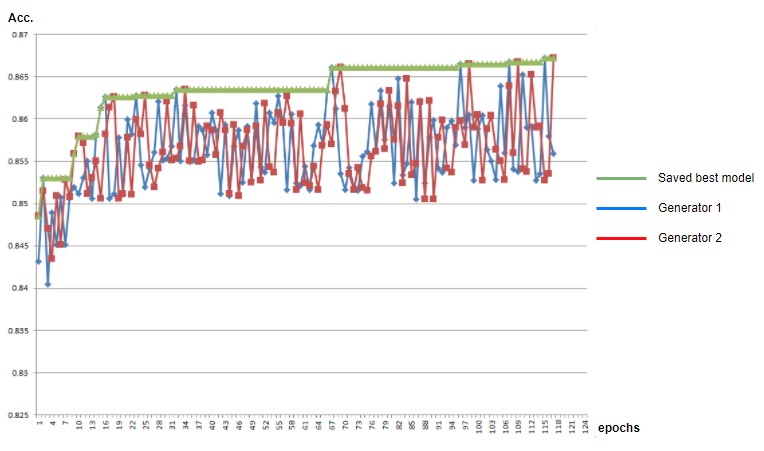}}
	\caption{Accuracy over number of epochs for the two generators on the ModelNet dataset.}
	\label{fig:modelnet result}
\end{figure}

The accuracy over number of epochs is shown in Fig.~\ref{fig:modelnet result}. The blue and red lines show the accuracies for Generator 1 and Generator 2, respectively. Green line is the saved model with the refined parameters providing the best accuracy. The accuracies of the original network (start accuracy) and the proposed approach (end accuracy) are presented in Table \ref{table:ModelNet_stop} for different early stopping criteria, more specifically, when $c$=1 and $c$=5. As can be seen, the proposed approach provides an increase in accuracy by autonomously and simultaneously refining nine parameters of the network in a systematic way.

\begin{table}[htb]
	\begin{center}
		\begin{tabular}{| c | c | c |}
			\hline
			Number of epochs (c) for early stopping  & 1    & 5  \\ \hline
			Start accuracy      & 83.43\%   & 84.31\%           \\ \hline
			\textbf{End accuracy}        & \textbf{85.71\%}    & \textbf{86.72\%}    \\ \hline
		\end{tabular}
	\end{center}
	\vspace{-0.1cm}
	\caption{Accuracies of the original network (start accuracy) and the proposed approach (end accuracy) with different early stopping criteria (when c=1, and c=5).} \vspace{-0.3cm}
	\label{table:ModelNet_stop}
	\vspace{-0.4cm}
\end{table}

\subsection{Experiments with UCI HAR Dataset and an LSTM-based network}
UCI HAR dataset~\cite{anguita2013public} is composed of Inertial Measurement Unit (IMU) data captured during activities of standing, sitting, laying, walking, walking upstairs and walking downstairs. These activities were performed by 30 subjects, and the 3-axial linear acceleration and 3-axial angular velocity were collected at a constant rate of 50Hz.

In this case, the network model shown in Fig.~\ref{fig:LSTM_evaluator} is used for evaluators. As can be seen, this network is an LSTM model. The output of each generator is a 9-dimensional vector which is composed of different parameter settings. More specifically, first four of the parameters are the number of neurons for two fully connected layers and two LSTM layers. Next four of the parameters indicate the choice of activation function for fully connected and two LSTM layers from (`Sigmoid', `Relu', `Linear', `Tanh') functions. Last of the nine parameters is a flag indicating whether to add a dropout layer between fully connected layers. In this case, $pm_{max}$ and $pm_{min}$ are set to be: $[4000, 4000, 2000, 2000, 1, 1, 1, 1, 1]$ and $[10, 10, 10, 10, 0, 0, 0, 0, 0]$, respectively.

\begin{figure}[h!]
	\centering
	\centerline{\includegraphics[width=.7\textwidth]{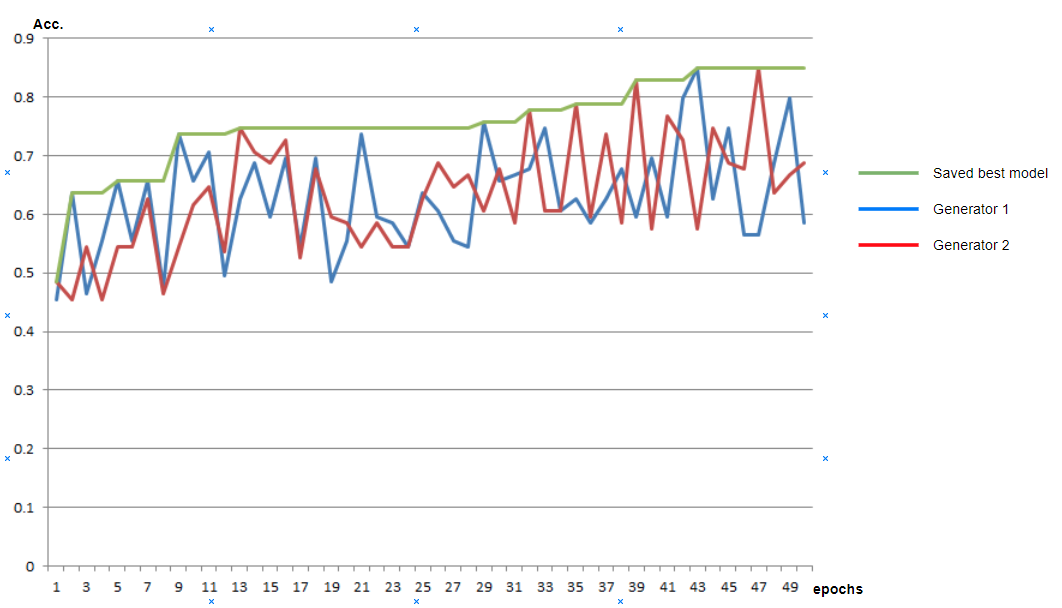}}
	\vspace{-0.2cm}
	\caption{Accuracy over number of epochs for the two generators on the UCI human activity recognition dataset.}
	\label{fig:lstm_acc}
	\vspace{-0.4cm}
\end{figure}

The accuracy over number of epochs is shown in Fig.~\ref{fig:lstm_acc}. The blue and red lines show the accuracies for Generator 1 and Generator 2, respectively. Green line is the saved model with the refined parameters providing the best accuracy. The accuracies of the original network (baseline accuracy) and the proposed approach are presented in the second row of Table \ref{table:all_scenarios}. As can be seen, the proposed approach provides an increase in accuracy for this LSTM network and this IMU dataset as well.

\begin{table}[htb]
	\begin{center}
		\begin{tabular}{| c | c | c |}
			\hline
			Dataset &  Baseline Accuracy    & Accuracy of the Proposed Method \\ \hline
			ModelNet      & 84.17\%   & \textbf{86.72\%}     \\ \hline
			UCI HAR      & 81.17\%   & \textbf{84.85\%}       \\ \hline
			Words built from Chars74k        & 85.5\%    & \textbf{86.64\%}    \\ \hline
		\end{tabular}
	\end{center}
	\vspace{-0.1cm}
	\caption{Accuracies of the original networks (baseline accuracy) and the proposed approach.}
	\vspace{-0.8cm}
	\label{table:all_scenarios}
\end{table}

\subsection{Experiments with Chars74k Dataset}
\begin{figure}[hbt!]
	\centering
	\begin{minipage}{.33\textwidth}
		\centering
		\includegraphics[width=0.8\linewidth]{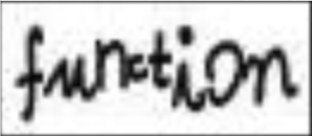}
	\end{minipage}%
	\begin{minipage}{.33\textwidth}
		\centering
		\includegraphics[width=0.8\linewidth]{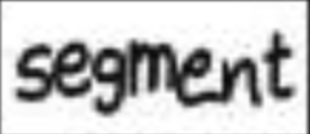}
	\end{minipage}
	\begin{minipage}{.33\textwidth}
		\centering
		\includegraphics[width=0.8\linewidth]{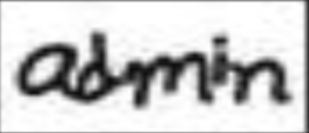}
	\end{minipage}
	\caption{Sample words built from the characters from the Chars74k dataset.}
	\label{fig:exp_words}
\end{figure}

We also tested our proposed approach with a word recognition method~\cite{li2016assisting}, which uses the characters from the Chars74k dataset~\cite{deCampos09} to build words. Chars74k dataset contains 64 classes (0-9, A-Z, a-z), 7705 characters obtained from natural images, 3410 hand-drawn characters using a tablet PC and 62992 synthesised characters from computer fonts giving a total of over 74K images. Some example words built from these characters are shown in Fig.~\ref{fig:exp_words}.

The work in \cite{li2016assisting} uses the network model shown in Fig.~\ref{fig:NLP} for character recognition, and then employs belief propagation for word recognition. In our experiments, we used the same network model in Fig.~\ref{fig:NLP} for our evaluators, and then performed the word recognition the same way to compare the word recognition accuracies. For the generators, the output is a 7-dimensional vector which is composed of different parameter settings. More specifically, first two of the parameters are the number of neurons for two fully connected layers. Next four of the parameters indicate the choice of activation function for fully connected and convolutional layers from (`Sigmoid', `Relu', `Linear', `Tanh') functions. The last of the seven parameters is a flag indicating whether to add a dropout layer between fully connected layers. In this case, $pm_{max}$ and $pm_{min}$ are set to be: $[4000, 4000, 4, 4, 4, 4, 1]$ and $[10, 10, 0, 0, 0, 0, 0]$, respectively.

The word recognition accuracies obtained by using the original network~\cite{li2016assisting} (baseline accuracy) and the proposed approach are presented in the last row of Table \ref{table:all_scenarios}. As can be seen, the proposed approach consistently provides an increase in accuracy for different types of networks and different datasets.

In Table \ref{table:parameters}, we present the parameters used in the original networks, and the parameters that were refined and optimized by the proposed method for all three different scenarios.

\begin{table}[htb]
	\begin{center}
		\resizebox{\columnwidth}{!}{%
			\begin{tabular}{| c | c | c |}
				\hline
				Dataset &  Baseline Parameters    & Parameters Chosen by the Proposed Method \\ \hline
				ModelNet      & [1024,256,1,1,1,1,1,1,0]  & [1242,1790,2,2,1,1,1,1,1]     \\ \hline
				UCI HAR      & [512,2014,1024,256,1,1,1,1,0]  & [771,939,2597,1403,1,1,1,1,0]       \\ \hline
				Words built from Chars74k  & [1024,256,1,1,1,1,0]  & [2804,2121,1,1,1,1,1]    \\ \hline
			\end{tabular}
		}
	\end{center}
	\vspace{-0.1cm}
	\caption{Parameters used by the original networks and the parameters that were refined and chosen by the proposed approach.}
	\vspace{-0.8cm}
	\label{table:parameters}
\end{table}

\section{Conclusion}\label{sec:concl}
In this paper, we have presented a novel and systematic method that autonomously and simultaneously optimizes multiple parameters of any given deep neural network by using a GAN-based approach. The set of parameters can include the number of neurons, the type of activation function, the choice of using drop out and so on. In our proposed approach, two different models compete and improve each other progressively with a GAN-based strategy. This approach can be used to refine parameters of different network architectures. Without loss of generality, the proposed method has been tested with three different neural network architectures, and three different datasets. The results show that the presented approach can simultaneously and successfully optimize multiple neural network parameters, and achieve increased accuracy in all three scenarios.

\clearpage

\bibliographystyle{bibstyle}
\bibliography{egbib}
\end{document}